\documentclass[10pt,twocolumn,letterpaper]{article}

\usepackage{iccv}
\usepackage{times}
\usepackage{epsfig}
\usepackage{graphicx}
\usepackage{amsmath}
\usepackage{amssymb}
\usepackage{booktabs}
\usepackage{multirow}
\usepackage[ruled]{algorithm2e}
\usepackage[accsupp]{axessibility}  

\usepackage[pagebackref=true,breaklinks=true,letterpaper=true,colorlinks,bookmarks=false]{hyperref}

\iccvfinalcopy 

\def\iccvPaperID{7359} 

\ificcvfinal\fi

\begin{document}
	\newcommand*\samethanks[1][\value{footnote}]{\footnotemark[#1]}
	\title{Video Instance Segmentation with a Propose-Reduce Paradigm}
	
	\author{
	Huaijia Lin$^1$\thanks{Equal Contribution.} \quad Ruizheng Wu$^1$\samethanks \quad Shu Liu$^{2}$ \quad Jiangbo Lu$^{2}$ \quad Jiaya Jia$^{1,2}$  \\
	$^1$The Chinese University of Hong Kong \quad $^2$SmartMore
	\\
	\texttt{\footnotesize \{linhj, rzwu, leojia\}@cse.cuhk.edu.hk \qquad \{sliu, jiangbo\}@smartmore.com }
	}
	
	
	\maketitle
	\ificcvfinal\thispagestyle{empty}\fi
	
	\begin{abstract}
		Video instance segmentation (VIS) aims to segment and associate all instances of predefined classes for each frame in videos. 
		Prior methods usually obtain segmentation for a frame or clip first, and merge the incomplete results by tracking or matching. These methods may cause error accumulation in the merging step. 
		Contrarily, we propose a new paradigm -- Propose-Reduce, to generate complete sequences for input videos by a single step.  
		We further build a sequence propagation head on the existing image-level instance segmentation network for long-term propagation.
		To ensure robustness and high recall of our proposed framework, multiple sequences are proposed where redundant sequences of the same instance are reduced. We achieve state-of-the-art performance on two representative benchmark datasets -- we obtain 47.6\% in terms of ${AP}$ on YouTube-VIS validation set and 70.4\% for ${J\&F}$ on DAVIS-UVOS validation set. Code is available at \url{https://github.com/dvlab-research/ProposeReduce}.
	\end{abstract}
	
	\section{Introduction}\label{Sec:intro}
	Video instance segmentation (VIS), proposed in \cite{yang2019vis}, is a task to segment all instances of the predefined classes in each frame. Segmented instances are linked in the entire video. It is important in the field of video understanding, and can be applied to video editing, autonomous driving, \etc. 
	Unlike image-level instance segmentation, VIS requires not only detection and segmentation of each frame, but also tracking of objects in the video, which make it a very challenging task.
	
	\begin{figure}[t]
		\begin{center}
			\includegraphics[width=1.0\linewidth]{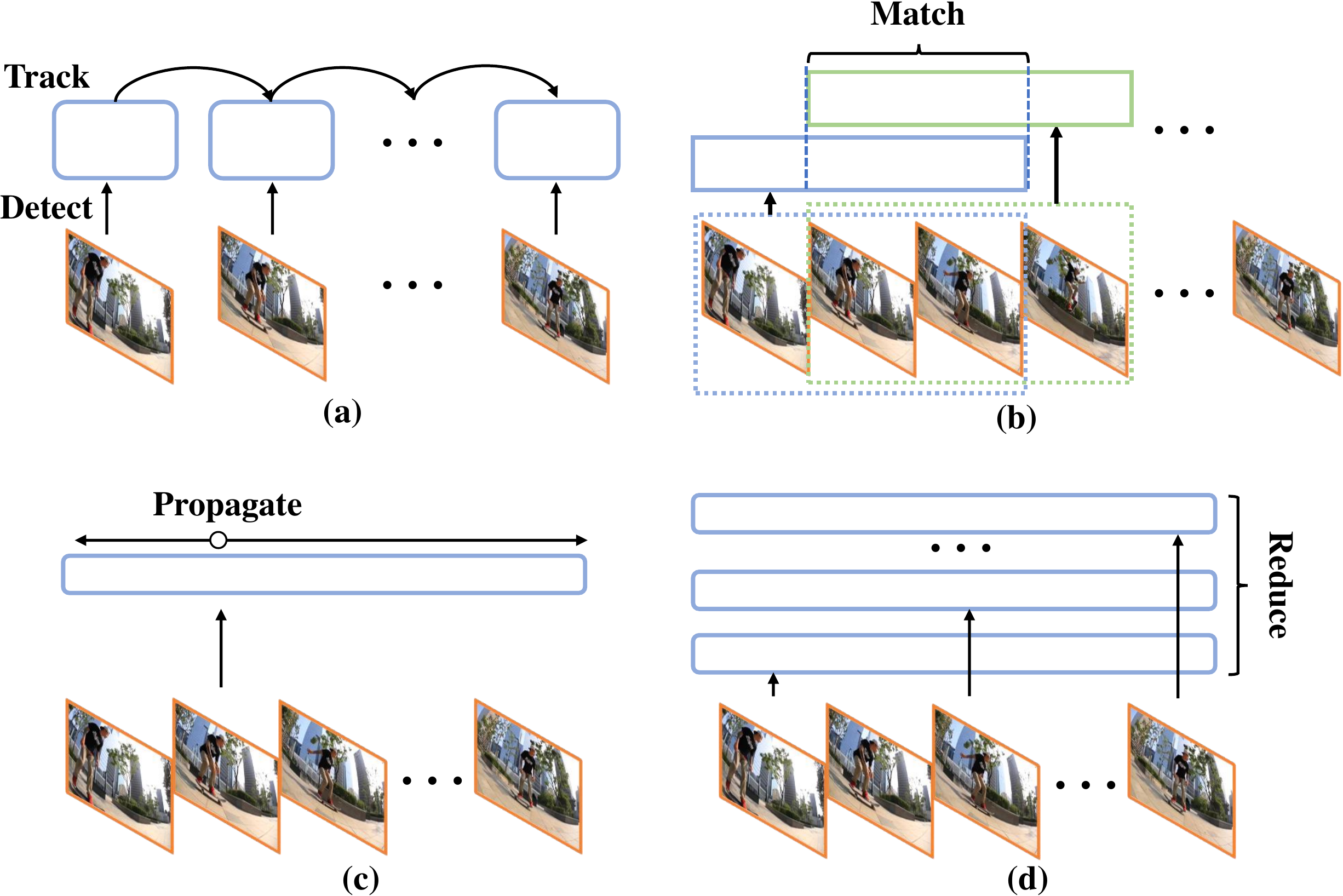}
		\end{center}
		\caption{Four paradigms of generating instance sequences in VIS.
			(a) \textbf{Track-by-Detect} links detected instances via frame-by-frame tracking.
			(b) \textbf{Clip-Match} matches overlapped sub-sequences between video-clips.
			(c) \textbf{An alternative} propagates detected instances from one key frame to the rest of a video.
			(d) Our proposed paradigm, named \textbf{Propose-Reduce}, generates instance sequence proposals based on multiple key frames and reduces redundant sequences of the same instances.} 
		\label{fig:paradigm}
	\end{figure}
	
	Recently, several approaches were proposed for this task~\cite{yang2019vis,cao2020sipmask,athar2020stem,bertasius2020maskprop,luiten2020unovost}. Based on the patterns of generating instance sequences, existing frameworks can be roughly categorized into two paradigms: `Track-by-Detect' (Fig.~\ref{fig:paradigm}(a))  and `Clip-Match' (Fig.~\ref{fig:paradigm}(b)). 
	The `Track-by-Detect' paradigm detects and segments instances for each individual frame, followed by obtaining instance sequences with frame-by-frame tracking~\cite{yang2019vis,cao2020sipmask}. 
	Differently, `Clip-Match' adopts the divide-and-conquer strategy. It divides an entire video into multiple short overlapped clips, and obtains VIS results for each clip and generates instance sequences with clip-by-clip matching~\cite{bertasius2020maskprop,athar2020stem}.
	Both of the paradigms need two independent steps to generate a complete sequence.
	They both generate multiple incomplete sequences (\ie, frames or clips) from a video, and merge (or complete) them by tracking/matching at the second stage. Intuitively, these paradigms are vulnerable to error accumulation in the process of merging sequences, especially when occlusion or fast motion exists.
	
	\begin{figure}[t]
		\begin{center}
			\includegraphics[width=1.0\linewidth]{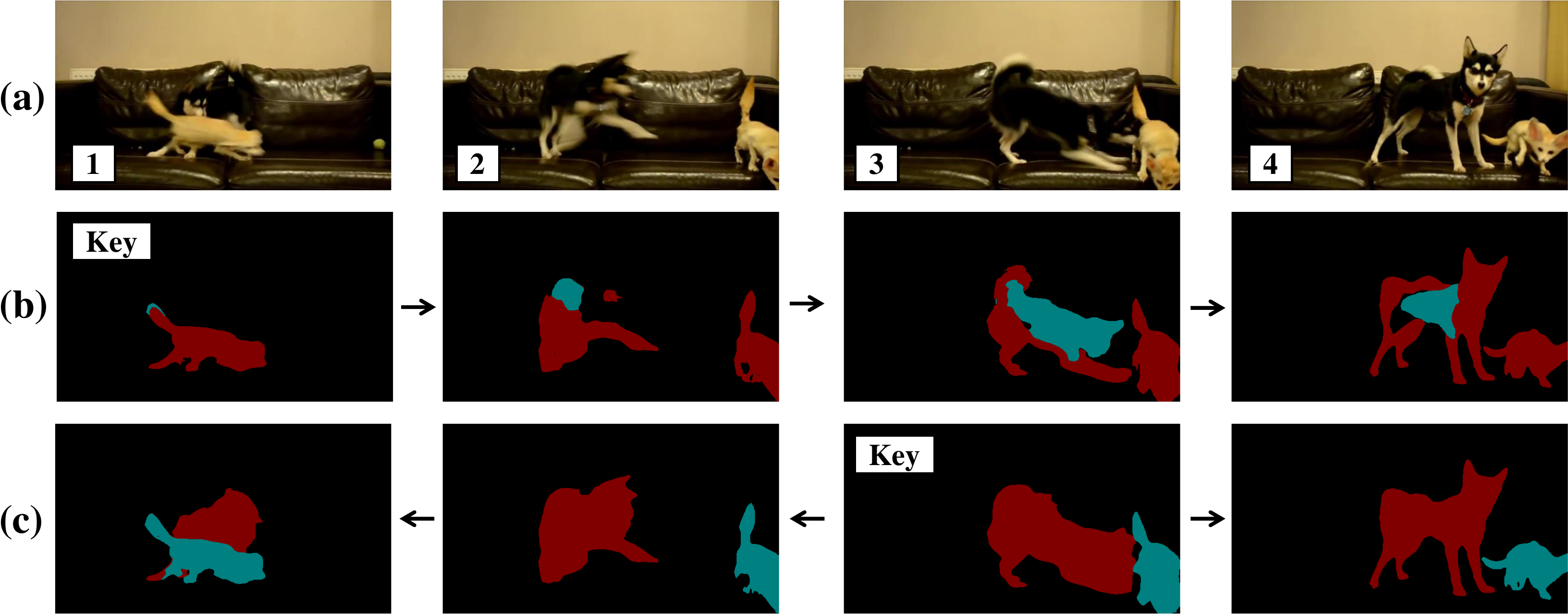}
		\end{center}
		\caption{Effect of propagation from different key frames.
			(a) Four ordered frames where severe occlusion occurs in the first frame.
			(b) Propagating from the first frame leads to inaccurate segmentation results caused by error accumulation.
			(c) Propagating from the third frame produces satisfactory segmentation masks, due to reasonable segmentation results in the key frame.
		}
		\label{fig:key}
	\end{figure}
	
	To avoid error accumulation brought by merging incomplete sequences, one intuitive solution is to generate a complete instance sequence for an entire video with only one step. As shown in Fig.~\ref{fig:paradigm}(c), starting from any key frame of a video, we can obtain instance sequences by propagating the instance segmentation results from this frame to all others. However, the propagation quality from different starting key frames varies a lot (as shown in Fig.~\ref{fig:key}). 
	One key frame may only contain part of the instances in a video, inappropriate for the whole sequence.
	
	For robust propagation and high recall to cover enough instances, we propose a new paradigm, named \textit{Propose-Reduce} (Fig.~\ref{fig:paradigm}(d)). It first produces sequence proposals from multiple key frames and reduces the redundant sequence proposals of the same instances. This paradigm not only discards the step of merging incomplete sequences, but also achieves robust results considering multiple key frames. 
	
	The idea behind Propose-Reduce is proved effective in the task of image-level object detection. 
	Methods for this task can be classified as one-stage~\cite{redmon2016you,liu2016ssd} and two-stage~\cite{ren2015faster,lin2017feature} detection frameworks. Compared with the one-stage frameworks, the two-stage ones first generate a large number of candidate proposals by the region proposal network (RPN)~\cite{ren2015faster} to ensure high recall, and then reduce abundant proposals by non-maximum suppression (NMS) as post-processing. 
	The great performance of the two-stage detection frameworks shows the potential of our {Propose-Reduce} in the video domain.

	Based on the above analysis, to propagate instance segmentation from each key frame to all other frames, 
	we design an additional module for long-term propagation, since a frame to be propagated can be far from the key frame.
	We propose attaching a sequence propagation head ({\it Seq-Prop head}) upon a widely-used image-level instance segmentation network: Mask R-CNN~\cite{he2017mask}. It enables sharing backbone features for multiple heads of different functions of 
	classification head, bounding box head, mask head and sequence propagation head. With the sharing backbone, our propagation module is light-weighted. 
	
	Besides, we adopt a memory propagation strategy on every key frame to enable long-term propagation.
	After obtaining sequence proposals from all key frames, we implement a variant of NMS to reduce redundant proposals at the sequence level.
	With the above design, our overall framework is neat and can be trained in an end-to-end fashion.
	The overall contributions are summarized below.
	\begin{itemize}
		\item We propose a new paradigm -- {\it Propose-Reduce}, for the task of video instance segmentation. This paradigm ensures high recall and does not require error-accumulating tracking/matching modules.
		\item Based on the paradigm, we propose a variant of Mask R-CNN for videos, named {\it Seq Mask R-CNN}. By adding an extra sequence propagation head upon Mask R-CNN, temporal relation is established across frames. 
		\item Our framework achieves new state-of-the-art results on YouTube-VIS~\cite{yang2019vis} validation set with 47.6\% in $\mathcal{AP}$, as well as DAVIS-UVOS~\cite{Caelles_arXiv_2019} validation set with {J\&F} score 70.4 \%.
		
	\end{itemize}
	
	\section{Related Work}
	\paragraph{Image-Level Instance Segmentation}
	Image-level instance segmentation is a classical computer vision task with many solutions~\cite{he2017mask,huang2019mask,liu2018path,wang2020solo,chen2019tensormask,xie2020polarmasK,bolya2019yolact,qi2020pointins} proposed. 
	They can be mainly divided into top-down~\cite{he2017mask,huang2019mask,liu2018path,bolya2019yolact,lee2020centermask}, bottom-up~\cite{liu2017sgn,newell2017associative}, and direct segmentation methods~\cite{wang2020solo,xie2020polarmasK}. 
	Among these methods, top-down structure is popular for high performance.
	It first utilizes detectors to detect objects and then segments them based on detected bounding boxes. 
	
	One representative method is Mask R-CNN~\cite{he2017mask}. It is built on a two-stage detector~\cite{ren2015faster}, adds a mask head for segmentation upon the detector, and keeps the original classification head as well as the bounding box head in the detector. 
	We design our framework based on Mask R-CNN from image to video domain. We introduce an extra sequence propagation head as well as a new paradigm for both spatial and temporal processing in the video instance segmentation task. Our method is simple and surprisingly effective.
	
	\begin{figure*}[t]
		\begin{center}
			\includegraphics[width=1.0\linewidth]{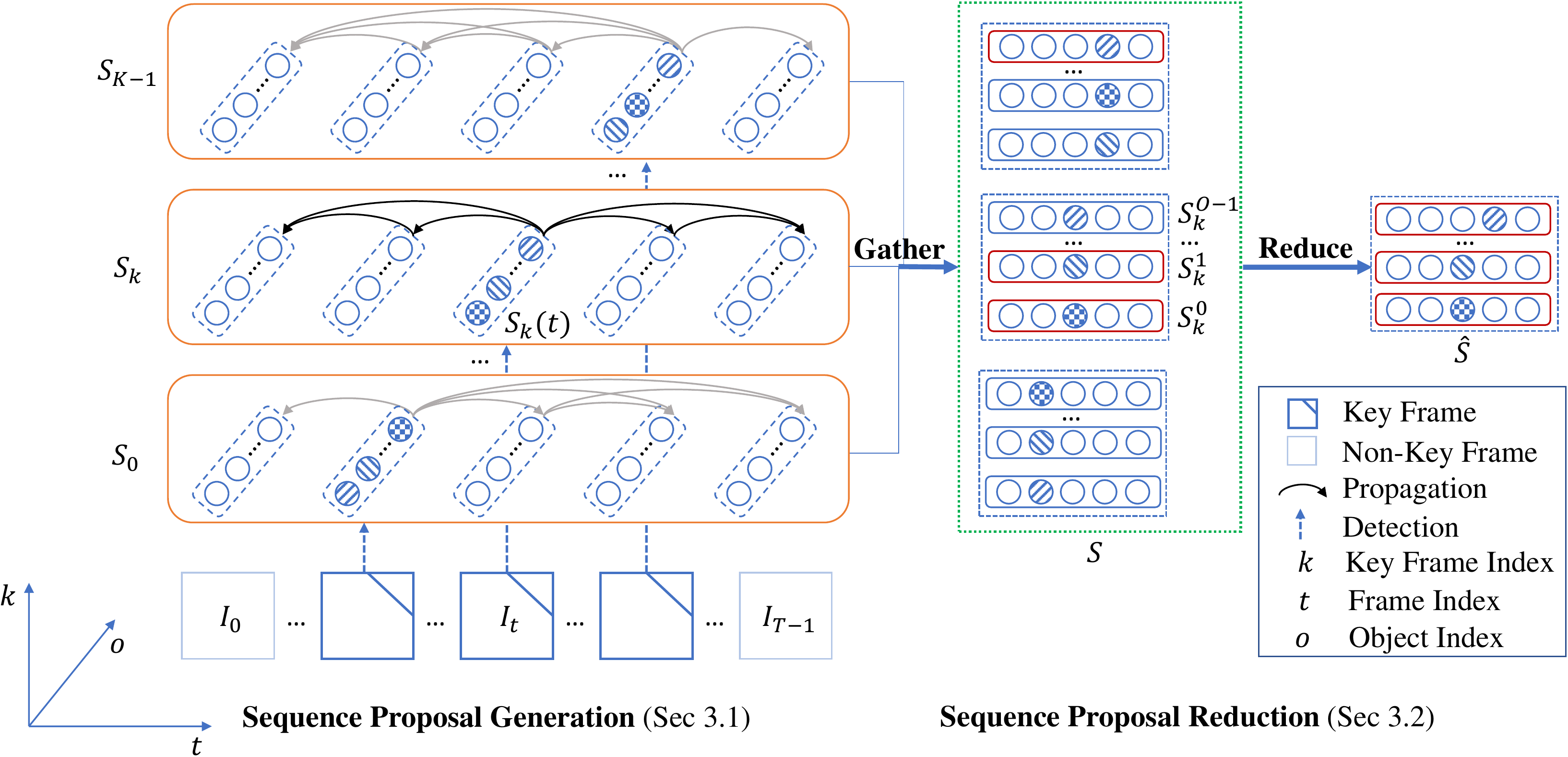}
		\end{center}
		\caption{The \textbf{Propose-Reduce} paradigm consists of two stages. 
			In the \textbf{Sequence Proposal Generation} stage, a sequence set $S_k$ is generated by first detecting $O$ instances at the $k^{th}$ key frame. We assume frame $t$ selected as the $k^{th}$ key frame for convenience. Instance set $S_k(t)$ at frame $t$ are then propagated to the whole video with \textit{memory $K$-Propagation} (Sec. 3.2.2). $K\times O$ sequences $\{S_k^o\}$ are gathered to form a redundant set $S$, which is reduced to the final sequence set $\widehat{S}$ in the \textbf{Sequence Proposal Reduction} stage. Different texture in circles differentiates among instances.}
		\label{fig:framework_a}
	\end{figure*}
	
	\paragraph{Video Instance Segmentation}
	Video instance segmentation (VIS) was introduced in \cite{yang2019vis}, which requires classification, segmentation and tracking on instances simultaneously in a video. existing work~\cite{yang2019vis,cao2020sipmask,athar2020stem,bertasius2020maskprop,luiten2020unovost,lin2020video} can be divided into two types based on the way of sequence generation.
	
	One straight-forward paradigm is `Track-by-Detect'~\cite{yang2019vis,cao2020sipmask,luiten2020unovost} with two parts: detection and tracking. In the detection part, instances are located with existing image-level instance segmentation methods~\cite{he2017mask} in a frame-by-frame manner. The detected instances are associated among different frames in the tracking part.
	Another paradigm is summarized as `Clip-Match' based methods~\cite{athar2020stem,bertasius2020maskprop}. It divides an entire video into multiple short clips and completes the VIS task in a clip-by-clip manner via propagation~\cite{bertasius2020maskprop} or spatial-temporal embedding~\cite{athar2020stem}. Neighboring clips are merged with matching (\eg, bipartite graph matching). 
	
	\vspace{-0.05in}
	
	\paragraph{Semi-supervised Video Object Segmentation}
	Semi-supervised video object segmentation (VOS) \cite{Perazzi2016, Pont-Tuset_arXiv_2017} refers to the problem of segmenting specified objects in videos given the annotated first frame. Research \cite{caelles2017one,perazzi2017learning,oh2019video,wug2018fast,voigtlaender2019feelvos,lin2019agss,goutam2020learning,lu2020video} was extensive. The most related methods to our framework are propagation-based, which propagate segmentation masks from the annotated first frame to the rest of the videos. Early research work~\cite{perazzi2017learning,khoreva2017lucid,yang2018efficient,wug2018fast} propagates in a frame-by-frame pipeline, which is fragile and easily fails in distant frames due to occlusion and fast motion. 
	
	Recent memory-based method STM~\cite{oh2019video} resolves the problem in long-term propagation. Several methods~\cite{wumemory,zhang2020transductive,lu2020video,seong2020kernelized} were proposed to improve the performance of STM. In this paper, the propagation strategy of our Seq-Prop head is inspired by STM. Compared with STM that adopts two separate backbones for extracting features, we make the Seq-Prop head light-weighted by sharing the same feature backbone with other heads in Mask R-CNN.
	
	\vspace{-0.05in}
	
	\paragraph{Unsupervised Video Object Segmentation}
	Compared with semi-supervised VOS, no annotated frame is given in unsupervised VOS (UVOS)~\cite{Caelles_arXiv_2019}. UVOS can be regarded as a variant of VIS, while VIS segments objects with pre-defined classes. UVOS is to segment class-agnostic salient objects.
	Recent work detects salient objects~\cite{wang2019learning,lu2020learning,song2018pyramid,zhou2020matnet}. Besides, 
	topological structures~\cite{ventura2019rvos,wang2019agnn} were proposed to obtain object segmentation in temporal domain. For example, RNN~\cite{ventura2019rvos} structure or graph convolution network~\cite{wang2019agnn} can be utilized. 
	Similar to VIS, detect-by-track \cite{luiten2020unovost} and clip-match based~\cite{athar2020stem} methods can be applied to UVOS, which first generate instance segmentation on a single frame (or clip) and track (or match) objects across frames or clips.
	Our paradigm can also be applied with the classification head modified from multi-class classification to two-class one (\ie, foreground and background).

	\section{Proposed Method}
	
	We propose the paradigm \textbf{Propose-Reduce} for the task of video instance segmentation. As shown in Fig.~\ref{fig:framework_a}, the paradigm consists of two stages. Redundant sequence proposals are generated in the first stage (Sec.~\ref{Sec:SPG}). We obtain instance segmentation on $K$ selected key frames (Sec.~\ref{Sec:KFS}). The segmentation results are then propagated to the whole video (Sec.~\ref{Sec:MKP}) with our proposed \textbf{Seq Mask R-CNN} framework (Sec.~\ref{Sec:SPH}).
	To reduce the redundancy in sequence proposals, in the second stage (Sec.~\ref{Sec:SPR}), a sequence reduction method is applied to all sequences for final sequence set as output.

	\subsection{Sequence Proposals Generation}\label{Sec:SPG}
	\subsubsection{Key Frames Selection}\label{Sec:KFS}
	To generate sequence proposals, we first select $K$ key frames to obtain their image-level instance segmentation masks. Specifically, for a $T$-frame video, the $K$ key frames $\{I_{g(0)},I_{g(1)},...,I_{g(K-1)}\}$ are selected at fixed intervals evenly, given by
	\begin{equation}
	\label{eq:keyframe}
	g(k) = \max\{\lfloor T/K \rfloor,~1\}\times k,~~k=0,..., K-1.
	\vspace{-0.1in}
	\end{equation}
	The number of key frames plays an important role in our design. 
	As described in Sec.~\ref{Sec:intro}, when selecting only one key frame, it degrades to paradigm (c) in Fig.~\ref{fig:paradigm}, where the final results are highly dependent of the instance segmentation quality in the selected key frames.
	However, when many key frames are selected, the computational cost of detection and propagation would increase.
	We accordingly choose a small number of key frames in our experiments. 
	
	For each key frame, we generate its instance segmentation results by multiple heads (\ie, bbox, classification and mask head) in Seq Mask R-CNN. For non-key frames, we only extract the backbone and FPN~\cite{lin2017feature} features of these frames for the following propagation step.
	It saves computation. 
	
	\begin{algorithm}[t]\label{alg:prop} 
		\caption{Memory $K$-Propagation} 
		\KwIn{Video frames $\{I_t |~t = 0,...,T-1 \}$, \\
			~~~~~~~~~~~~Key frames number $K$. }
		\KwOut{Instance sequence proposal set $S$. }
		\For{$k=0; k \textless K; k \leftarrow k+1$}
		{
			
			$t = g(k)$; ~\textcolor[rgb]{0.5,0.5,0.5}{\footnotesize // (Eq.~\ref{eq:keyframe})} \\  
			$S_k(t) \leftarrow \text{Detect}(I_t)$\;
			\textcolor[rgb]{0.5,0.5,0.5}{\footnotesize /* Forward Direction */} \\
			$\mathcal{M} \leftarrow \{S_k(t)\}$\;    
			\For{$i = t + 1; i \textless T; i \leftarrow i + 1$}
			{
				$S_k(i) \leftarrow \text{Propagate}(\mathcal{M}, I_i)$; ~\textcolor[rgb]{0.5,0.5,0.5}{\footnotesize // (Sec.~\ref{Sec:SPH}) } \\
				$\mathcal{M} \leftarrow \mathcal{M} \cup S_k(i)$\;
			}
			\textcolor[rgb]{0.5,0.5,0.5}{\footnotesize /* Backward Direction */} \\
			$\mathcal{M} \leftarrow \{S_k(t)\}$\;    
			\For{$j = t - 1; j \ge 0; j \leftarrow j - 1$}
			{
				$S_k(j) \leftarrow \text{Propagate}(\mathcal{M}, I_j)$; ~\textcolor[rgb]{0.5,0.5,0.5}{\footnotesize // (Sec.~\ref{Sec:SPH})} \\
				$\mathcal{M} \leftarrow \mathcal{M} \cup S_k(j)$\;
			}
			$S_k \leftarrow (S_k(0), S_k(1), ..., S_k(T-1))$\;
			$k \leftarrow k + 1$\; 
		}
		$S \leftarrow S_0 \cup S_1 \cup ... \cup S_{K-1}$; ~~~~~~\textcolor[rgb]{0.5,0.5,0.5}{\footnotesize // Gather }  \\
		return $S$\;	
		
	\end{algorithm}
	
	\subsubsection{Memory $K$-Propagation}\label{Sec:MKP}
	The instance masks on $K$ selected key frames are propagated bi-directionally to obtain $K$ sets of mask sequences, \ie, $\{S_0, S_1, ..., S_{K-1}\}$, as illustrated in Alg.~\ref{alg:prop}.  After all propagation finished, we gather $K$ sequence sets from different key frames into one set $S = S_0 \cup S_1 \cup ... \cup S_{K-1}$.
	
	As shown in Alg.~\ref{alg:prop}, we maintain a memory $\mathcal{M}$ to alleviate error accumulation in long-term propagation \cite{oh2019video}. It stores the encoded feature of previously segmented frames and propagates mask information to the current frame.
	The operations on $\mathcal{M}$ (\eg, read and update) are similar to that of STM~\cite{oh2019video}. The difference is that the memory in \cite{oh2019video} is utilized to propagate from the annotated first frame to the end of a video, while our work propagates the estimated mask from a key frame to the beginning and end of the video for $K$ times. 
	
	Directly applying STM to our paradigm requires another two backbones to extract features for memory and query. 
	Instead, we design an additional propagation module that can be seamlessly inserted into the image-level instance segmentation frameworks.
	
	\begin{figure*}[t]
		\begin{center}
			\includegraphics[width=1.0\linewidth]{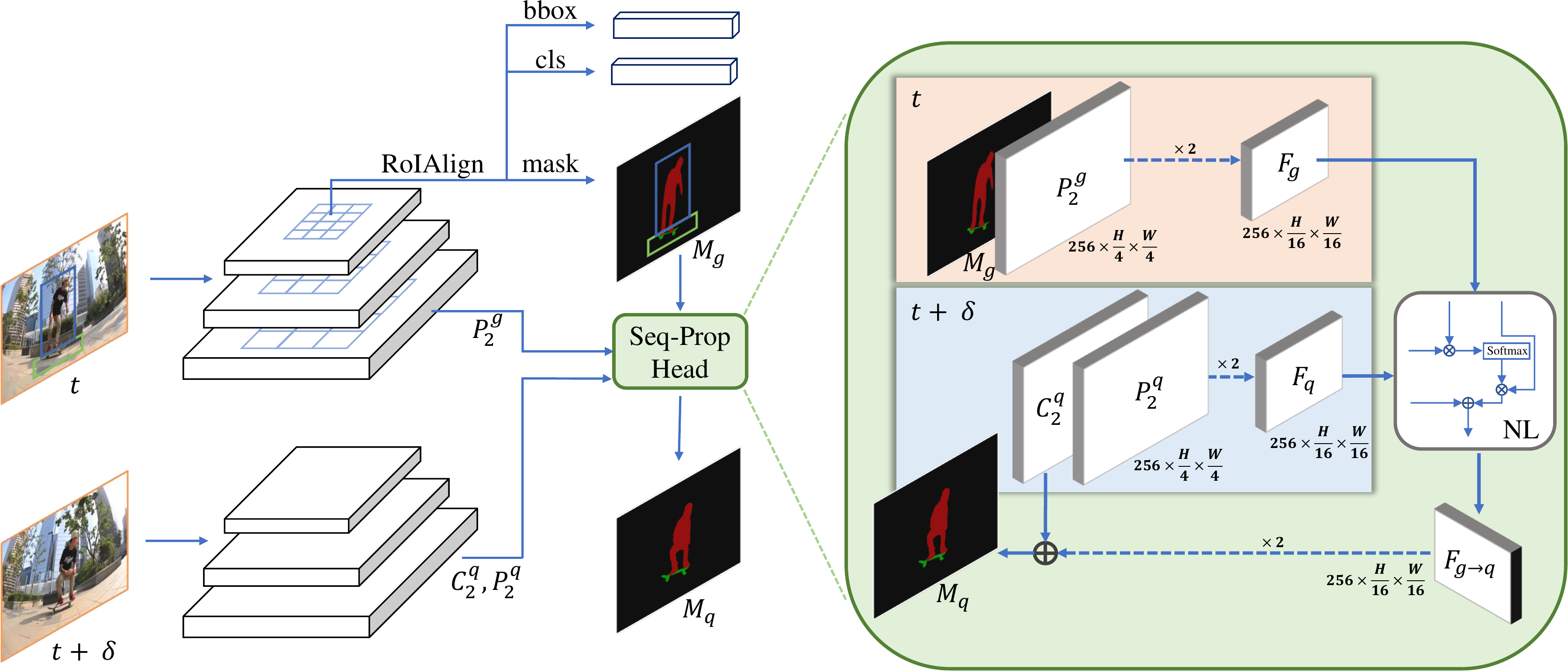}
		\end{center}
		\caption{Framework of \textbf{Seq Mask R-CNN}. We adopt \textbf{Seq-Prop head} on Mask R-CNN for propagating instance masks from the guidance frame at time $t$ to a query frame at time $t+\delta$. $P_2^g$, $P_2^q$ are the largest FPN~\cite{lin2017feature} feature of input images, and $C_2^q$ is the largest backbone feature. \textbf{NL} is a non-local operation~\cite{wang2018non}. `$\times 2$' denotes $2$ consecutive residual blocks. `$\otimes$' and `$\oplus$' denote matrix multiplication and summation, respectively. Detailed architectures are illustrated in the supplementary files. }
		\label{fig:framework}
		\vspace{-0.1in}
	\end{figure*}

	\subsubsection{Sequence Mask R-CNN}\label{Sec:SPH}
	We incorporate a propogation head (Seq-Prop head) on the top of Mask R-CNN for memory $K$-Propagation, which is called Sequence Mask R-CNN (Seq Mask R-CNN).
	
	\vspace{-0.1in}
	
	\paragraph{Architecture} The architecture of Seq Mask R-CNN is shown in Fig.~\ref{fig:framework}. It is based on outputting instance segmentation results for a single image, to which we add an extra propagation head that propagates instance masks to other frames.
	
	Fig.~\ref{fig:framework} illustrates an example that takes two frames as input. We refer to them as the guidance frame (the $t^{th}$ frame) and the query frame (the $(t + \delta)^{th}$ frame). 
	For the guidance frame, we take the estimated mask $M_g$ and largest FPN~\cite{lin2017feature} feature $P_2^g$ as input for encoding feature $F_g$. For a query frame, we take its $P_2^q$ feature as input to obtain feature $F_q$.
	
	With the two encoded feature maps, we utilize a non-local operation (NL)~\cite{wang2018non} to propagate mask information from the guidance frame to the query one and obtain the propagated feature $F_{g \to q}$.
	Finally, $F_{g \to q}$ and the largest backbone feature $C_2^q$ from the query frame are utilized for decoding and generating the query mask $M_q$. 
	The FPN feature $P_2^q$ and the backbone feature $C_2^q$ are utilized in encoding and decoding respectively, since they contain the richest semantic and detailed information on multiple instances.
	
	\vspace{-0.1in}
	
	\paragraph{Training}
	In the training stage, we randomly select two frames as input for memory efficiency, \ie, one guidance and one query frame. 
	In one epoch, the query frame in a pair of frames is selected once per frame per video. The guidance frame is randomly sampled from the same video.
	Besides, in order to make the Seq-Prop head learn to propagate from the imperfect segmented masks, we utilize the estimated instance masks instead of the ground-truth ones as the guidance input for training. It makes the head more robust at the inference stage.
	
	To train our overall framework, we adopt a multi-task loss $\mathcal{L}=\mathcal{L}_{cls}+\mathcal{L}_{box}+\mathcal{L}_{mask}+\mathcal{L}_{prop}$. The classification loss $\mathcal{L}_{cls}$, bounding-box loss $\mathcal{L}_{box}$, and mask loss $\mathcal{L}_{mask}$ are the same as those in Mask R-CNN~\cite{he2017mask}. 
	As for the propagation loss $\mathcal{L}_{prop}$ utilized to train Seq-Prop head, we adopt a 
	scale-balanced soft IoU loss~\cite{lin2019agss}, since the Seq-Prop head propagates multi-scale instances masks at the same time.
	
	\vspace{-0.1in}
	
	\paragraph{Inference}
	During the stage of inference, the guidance frame input is replaced by a memory pool (illustrated in Alg.~\ref{alg:prop}), which stores the encoded features from the frames that have been propagated. Specifically, for each iteration of propagation, memory is updated by appending the encoded feature of the current frame, which increases model's robustness to occluded instances~\cite{oh2019video}.
	
	By sharing backbone features with the other three heads in Seq Mask R-CNN, our propagation head discards two heavy encoders for memory and query in STM~\cite{oh2019video}.

	\subsection{Sequence Proposals Reduction}\label{Sec:SPR}
	Redundant sequence proposals exist after the first stage where sequences of the same instance may be generated for multiple times from different key frames. To reduce redundancy, inspired by NMS~\cite{girshick2014rich,girshick2015fast,ren2015faster,rothe2014non} that is widely used in image-level instance segmentation in post-processing, we adopt a variant of NMS for sequences reduction. To apply it to sequences, three key elements in NMS need to be defined, i.e., the input sequence sets, sequence score and sequences IoU.
	
	\vspace{-0.1in}
	\paragraph{Input Sequence Set}
	In the stage of sequence proposal generation, we obtain $K$ sets of sequence proposals $\{S_0, S_1, ..., S_{K-1}\}$ gathered as $S$. For each sequence set $S_k$, we have its corresponding mask $M(S_k)$ and classification score $C(S_k)$. 
	We set the maximum instance number in a key frame as $O$. Then $S_k$ can be represented as a set of instance sequences $\{S_k^o\}$, where $o \in [0, O$-$1]$. Correspondingly, their masks and scores are defined as the set of $\{M(S_k^o)\}$ and $\{C(S_k^o)\}$,
	where $M(S_k^o) \in \{0,1\}^{T\times H\times W}$ and the score of each sequence $C(S_k^o)$ is defined later.
	
	Accordingly, we obtain the input sequence set as $S=\{S_k^o\}$, where $k \in [0, K$-$1], o \in [0, O$-$1]$, which consists of a maximum of $K \times O$ instance sequences. Since the instance number per key frame is less than $O$ in most cases, many sequences in $S$ are empty and the sequence number is much less than $K \times O$. 
	Our target is to reduce the redundant sequence set $S$ into a final sequence set $\widehat{S}$.
	
	\vspace{-0.1in}
	\paragraph{Sequence Score}
	The score for each instance sequence $C(S_k^o)$ reflects its priority to be selected. 
	To represent the priority of an instance sequence, we consider all frames in this sequence. For each instance, on any frame $I_t$, we obtain its classification score $C(S_k^o(t)) \in [0,1]^{|\mathbb{C}|}$ from the classification head of Seq Mask R-CNN, where $|\mathbb{C}|$ indicates the number of instance classes. We average the scores among all frames and take the max score among $|\mathbb{C}|$ classes as the score of this instance sequence. The score for each sequence $C(S_k^o)$ is defined as 
	\begin{equation}\label{eq:seq_score}
	C(S_k^o) = \max_{|\mathbb{C}|} \frac{1}{T} \sum_{t=0}^{T-1} {C(S_k^o(t))} \;.
	\vspace{-0.1in}
	\end{equation}
	
	\vspace{-0.1in}
	\paragraph{Sequences IoU}
	Intersection-over-union (IoU) between two sequences measures their overlap. 
	We calculate the mask IoU instead of the bounding-box IoU to measure the overlap more precisely.
	We denote the masks of two sequences as $M(S_k^o)$ and $M(S_{\tilde{k}}^{\tilde{o}})$, where $M(S_k^o)$ indicates the mask sequence of the $o^{th}$ instance from the $k^{th}$ key frame, and $M(S_{\tilde{k}}^{\tilde{o}})$ is similarly defined.
	Then the $IoU$ between two sequences, \ie, $IoU(S_{k}^{o}$ and $S_{\tilde{k}}^{\tilde{o}})$, is computed as
	\begin{equation}\label{eq:seq_iou}
	IoU(S_{k}^{o}, S_{\tilde{k}}^{\tilde{o}}) = \frac{\sum_{t = 0}^{T-1} | M(S_k^o(t)) \cap M(S_{\tilde{k}}^{\tilde{o}}(t)) |}{\sum_{t = 0}^{T-1} | M(S_k^o(t)) \cup M(S_{\tilde{k}}^{\tilde{o}}(t)) |} \;,
	\end{equation}
	where $M(S_k^o(t))$ and $M(S_{\tilde{k}}^{\tilde{o}}(t))$ are the masks of the $t^{th}$ frame from the two sequences $S_k^o$ and $S_{\tilde{k}}^{\tilde{o}}$, respectively.
	
	With the defined sequence set, sequence score and sequences IoU, we directly apply the traditional NMS algorithm to the sequence set to reduce the redundant sequences. 
	More details of this algorithm is included in our supplementary files. 
	The sequence set $\widehat{S}$ after NMS is our final result for the task of VIS.

	\begin{table*}[t!]
		\centering
		\scalebox{0.95}{
			\begin{tabular}
				{cc|cc|c|cccc}
				\toprule[1pt]
				Paradigm & Method & Backbone & HR-Ref & $\mathcal{AP}$ & $\mathcal{AP}@50$ & $\mathcal{AP}@75$ & $\mathcal{AR}@1$ & $\mathcal{AR}@10$ \\ 
				\midrule[0.5pt]
				&	MaskTrack \cite{yang2019vis} & ResNet-50 & & 30.3 & 51.1 & 32.6 & 31.0 & 35.5 \\
				Track-by-Detect	&	SipMask \cite{cao2020sipmask} & ResNet-50 & & 33.7 & 54.1 & 35.8 & 35.4 & 40.1 \\
				&	EnsembleVIS \cite{luiten2019ensvis} & ResNeXt-101* & $\checkmark$ & 44.8 & - & 48.9 & 42.7 & 51.7 \\
				\midrule[0.5pt]
				&	STEm-Seg \cite{athar2020stem} & ResNet-50 & & 30.6 & 50.7 & 33.5 & 31.6 & 37.1 \\
				&	STEm-Seg \cite{athar2020stem} & ResNet-101 & & 34.6 & 55.8 & 37.9 & 34.4 & 41.6 \\
				Clip-Match	&	MaskProp \cite{bertasius2020maskprop} & ResNeXt-101 & $\checkmark$ & 44.3 & - & 48.3 & - & - \\
				&	MaskProp \cite{bertasius2020maskprop} & STSN~\cite{bertasius2018stsn}-ResNeXt-101 & & 44.7 & - & - & - & - \\
				&	MaskProp \cite{bertasius2020maskprop} & STSN~\cite{bertasius2018stsn}-ResNeXt-101 & $\checkmark$ & 46.6 & - & 51.2 & 44.0 & 52.6 \\
				\midrule[0.5pt]
				&	Ours & ResNet-50 & & 40.4 & 63.0 & 43.8 & 41.1 & 49.7 \\
				Propose-Reduce	&	Ours & ResNet-101 & & 43.8 & 65.5 & 47.4 & 43.0 & 53.2 \\
				&	Ours & ResNeXt-101 & & \textbf{47.6} & \textbf{71.6} & \textbf{51.8} & \textbf{46.3} & \textbf{56.0} \\
				\bottomrule[1pt]
		\end{tabular}}
		\vspace{0.05in}
		\caption{Quantitative results of video instance segmentation in YouTube-VIS validation set. `HR-Ref' indicates post-processing that resizes cropped masks to a large resolution and refines details with extra convolutional layers. *: EnsembleVIS adopts multiple models and their largest backbone is ResNeXt-101. 
		}
		\label{tab:exp_ytv}
	\end{table*}
	
	\section{Experiments}
	\subsection{Datasets} \label{Sec:dataset}
	\paragraph{YouTube-VIS~\cite{yang2019vis}} YouTube-VIS dataset is currently the largest dataset for video instance segmentation task. It contains 2,238 training videos and 302 validation videos, with 40 categories involved. Validation scores are evaluated on online benchmark. Similar to image instance segmentation~\cite{lin2014coco} , the benchmark adopts Average Precision ($\mathcal{AP}$) and Average Recall ($\mathcal{AR}$) metrics to evaluate the sequence accuracy averaged over the category set. 
	
	\vspace{-0.2in}
	\paragraph{DAVIS-UVOS~\cite{Caelles_arXiv_2019}} DAVIS-UVOS dataset is proposed for unsupervised video object segmentation for salient generic objects. It contains 60 training videos and 30 validation videos with high-quality annotations.
	This task can be viewed as a special case of video instance segmentation with 2 categories (foreground and background). 
	In the evaluation stage, it considers no more than $20$ predicted sequences in a video and measures the average between $\mathcal{J}$ scores (the mean IoU between the estimated mask and ground-truth) and $\mathcal{F}$ scores (the F-measure of the estimated mask boundaries) via bipartite graph matching.	
	
	\subsection{Implementation Details}\label{Sec:Imp}
	The training data in the above datasets is not sufficient, resulting in over-fitting. To alleviate this issue, we adopt 80K training images in the COCO~\cite{lin2014coco} dataset (image instance segmentation) for compensation (also adopted in ~\cite{athar2020stem}).
	For each image in COCO, we augment it with $\pm30^{\circ}$ rotation to generate a three-frame pseudo video. For the training on YouTube-VIS, we only select images with overlapping categories in COCO. For DAVIS-UVOS, we select all images from COCO and treat all annotated instances as one category, \ie, foreground. 
	
	Our training consists of two stages, \ie, \textbf{main-training stage} and \textbf{finetuning stage}. In the main-training stage, we first train our model on the mixed dataset including COCO and the video dataset (\ie, YouTube-VIS, DAVIS-UVOS) for 4 epochs with $640\times320$ input size. In the finetuning stage, for YouTube-VIS dataset, the model is trained on this dataset with the same input size, while the model finetuned on DAVIS-UVOS dataset takes $854\times480$ size as input. We finetune the models for 5 epochs for both datasets. 
	
	All models are trained with 6 NVIDIA Titan X GPUs, implemented by PyTorch. The training time takes about $2$-$4$ days for each dataset. We set $K$ as $6$ for YouTube-VIS and $4$ for DAVIS-UNVOS (see Sec.~\ref{sec:abl_keyframe}). 
	More details are included in our supplementary files.

	\begin{table}[t!]
		\centering
		\scalebox{0.875}{
			\begin{tabular}
				{c|c|ccc}
				\toprule[1.0pt]
				Method & Backbone & $\mathcal{J\&F}$ & $\mathcal{J}$-Mean & $\mathcal{F}$-Mean  \\ 
				\midrule[0.5pt]
				RVOS \cite{ventura2019rvos} & ResNet-101 & 41.2 & 36.8 & 45.7 \\
				STEm-Seg \cite{athar2020stem} & ResNet-101 & 64.7 & 61.5 & 67.8 \\
				UnOVOST \cite{luiten2020unovost} & ResNet-101* & 67.9 & 66.4 & 69.3 \\
				\midrule[0.5pt]		
				Ours & ResNet-101 & 68.3 & 65.0 & 71.6 \\ 
				Ours & ResNeXt-101 & \textbf{70.4} & \textbf{67.0} & \textbf{73.8} \\
				\bottomrule[1.0pt]
		\end{tabular}}
		\vspace{0.05in}
		\caption{Quantitative results of unsupervised video object segmentation on DAVIS-UVOS validation set. *: UnOVOST combines multiple models and their largest backbone is ResNet-101. }
		\vspace{-0.15in}
		\label{tab:exp_davis}
	\end{table}
	
	\subsection{Main Results} 
	
	\paragraph{YouTube-VIS} 
	The quantitative results on YouTube-VIS are included in Table~\ref{tab:exp_ytv}. We list the backbone \cite{he2016deep,xie2017aggregated,bertasius2018stsn} used in different methods for fair comparison. 
	MaskProp, the SOTA method, takes a strong backbone (\ie, STSN~\cite{bertasius2018stsn}-ResNeXt-101) to extract spatial-temporal features, and a stronger detection head (\ie, HTC \cite{chen2019htc}) to refine detection results iteratively. In contrast, our best model only uses ResNeXt-101 to extracting spatial representation features and the vanilla head in Mask R-CNN for detection. Our model already outperforms MaskProp by $1\%$ in terms of $\mathcal{AP}$ and $3.4\%$ in terms of $\mathcal{AR}@10$.
	
	The large improvement of recall stems from the sampling strategy on multiple key frames.
	Note that MaskProp adopts a post-process that refines masks to gain $1.9\%$ improvement, while Seq Mask R-CNN does not adopt this post-process. 
	The previous best method that only extracts spatial features is EnsembleVIS, which combines four separate networks into a complex system, including detection \cite{he2017mask}, classification \cite{xie2017aggregated}, re-identification \cite{luiten2020unovost}, and segmentation \cite{chen2018deeplabv3}. Our \textit{single-model} method surpasses EnsembleVIS by $2.8\%$ in $\mathcal{AP}$ and $4.3\%$ in $\mathcal{AR}@10$.
	
	\vspace{-0.1in}
	\paragraph{DAVIS-UVOS} 
	We also evaluate our approach on DAVIS-UVOS dataset, as shown in Table~\ref{tab:exp_davis}. 
	The SOTA method UnOVOST combines multiple models (\eg, Mask R-CNN \cite{he2017mask}, PWC-Net \cite{sun2018pwc} and ReID Net \cite{wu2019reidNet}) into a complex system. Our \textit{single-model} method with ResNet-101 backbone achieves a comparable performance. With a stronger backbone (\ie, ResNeXt-101), our method outperforms UnOVOST in both $\mathcal{J}$ and $\mathcal{F}$ scores. 
	Compared with SOTA single-model method STEm-Seg, our method with the same backbone (\ie, ResNet-101) surpasses it by $3.6\%$ in $\mathcal{J\&F}$. 
	
\begin{figure}[t]
		\begin{center}
			\scalebox{0.99}{
				\includegraphics[width=1.0\linewidth]{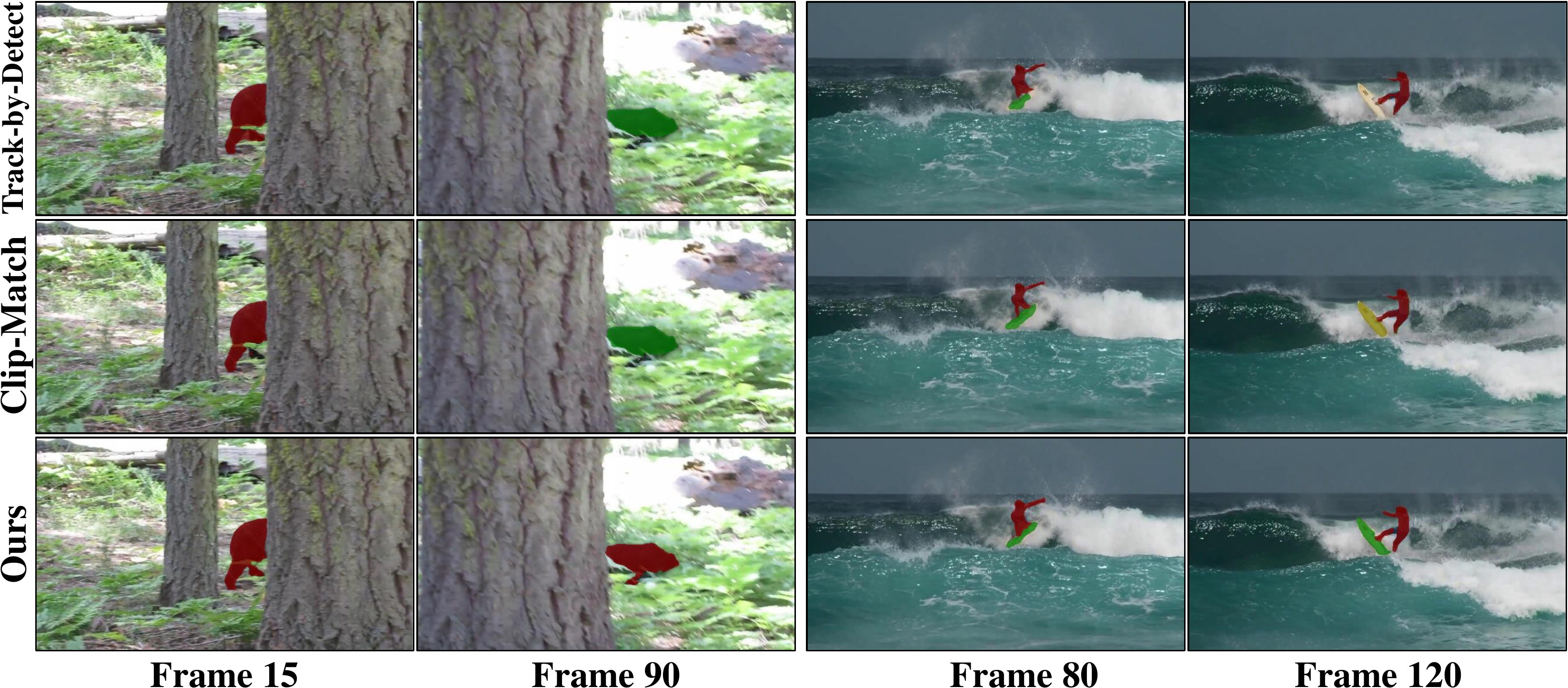}
			}
		\end{center}
		\vspace{-0.1in}
		\caption{Visual comparison of different paradigms on challenging scenarios. Frames are sampled before and after occlusion. }
		\label{fig:vis_paradigm}
	\end{figure}

	\vspace{-0.15in}
	\paragraph{Visualization} 
	We further present the comparison with previous paradigms (\cite{cao2020sipmask,athar2020stem})  in long-term occlusion scenarios, as shown in Fig.~\ref{fig:vis_paradigm}. Salient objects (bear/surfboard) are occluded by trees/waves in multiple frames. Track-by-Detect~\cite{cao2020sipmask} fails to re-identify the same instance with distorted appearance. Clip-Match \cite{athar2020stem} treats them as two instances due to being out of the matching scopes. In contrast, our paradigm generates complete sequences via long-term propagation. More visual results are shown in Fig.~\ref{fig:visu}. Our method fails to propagate masks with highly consistently-occluded instances of the same category (\ie, person).

	\begin{figure*}[t]
		\begin{center}
			\scalebox{0.99}{
				\includegraphics[width=1.0\linewidth]{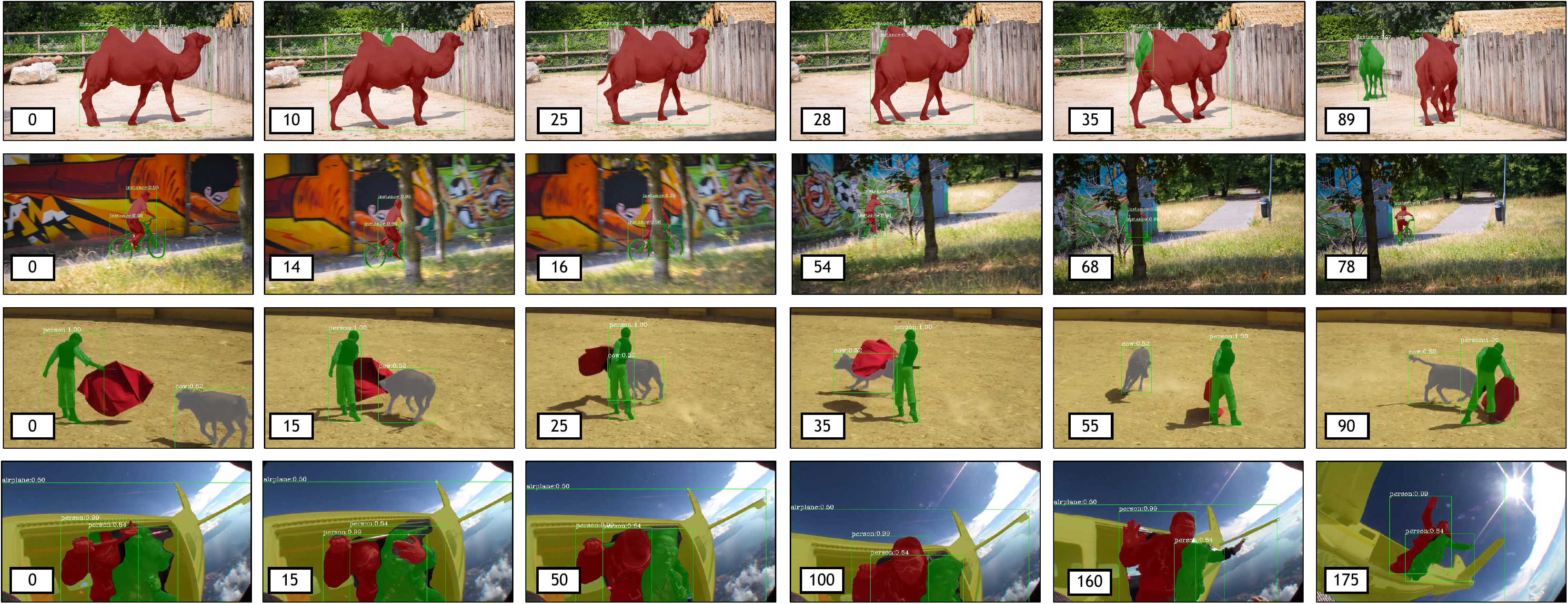}
			}
		\end{center}
		\vspace{-0.1in}
		\caption{Visual results on DAVIS-UVOS and YouTube-VOS. Frames are sampled at challenging moments (\eg, fast motion). We also show in the last row a failure case of overlapped same-category instances, where the arm of one occluded person is segmented to the other. Category `Instance' in DAVIS-UVOS denotes the salient generic object.}
		\label{fig:visu}
		\vspace{-0.05in}
	\end{figure*}
	
	\subsection{Ablation Experiments}\label{Sec:ablation}
	All the ablation experiments are conducted with the ResNeXt-101~\cite{xie2017aggregated} backbone on YouTube-VIS and DAVIS-UVOS validation sets.

	\begin{table}[t]
		\centering
		\scalebox{1}{
			\begin{tabular}
				{c|cc}
				\toprule[1.0pt]
				Variants & YouTube-VIS & DAVIS-UVOS \\
				\midrule[0.5pt]
				Main-training only & 46.2 & 67.3 \\
				Finetuning only & 40.8 & 48.9 \\
				\midrule[0.5pt]
				Both & \textbf{47.6} & \textbf{70.4} \\
				\bottomrule[1.0pt]
			\end{tabular}
		}
		\vspace{0.05in}
		\caption{Training data analysis on YouTube-VIS and DAVIS-UVOS validation set. `Both' denotes two-stage training, including main-training and finetuning. We report $\mathcal{AP}$ for YouTube-VIS and $\mathcal{J\&F}$ for DAVIS-UVOS.
		}\label{tab: abl_data}
		\vspace{-0.1in}
	\end{table}
	
	\vspace{-0.15in}
	\paragraph{Training Stage}\label{Sec: trainstage}
	We conduct experiments to study the effect of different training stages (Sec.~\ref{Sec:Imp}), as shown in Table~\ref{tab: abl_data}. With the finetuning stage only, the large drop of performance indicates that insufficient video data leads to over-fitting.
	Adopting the main-training stage only alleviates over-fitting. It is still hard to reach the performance by two-stage training, since there exist a domain gap between image (\ie, COCO) and video datasets (\ie, YouTube-VIS, DAVIS-UVOS).
	
	\vspace{-0.15in}
	\paragraph{Sequence Reduction}
	Tab.~\ref{tab: abl_nms} reports the ablation results with and without sequence reduction. Its effects on YouTube-VIS and DAVIS-UVOS are different for their evalution metric. The evaluation metric in YouTube-VIS is sensitive to false positive. Sequence reduction significantly increases $\mathcal{AP}$ at the cost of a slight decrease of $\mathcal{AR}$. For DAVIS that does not penalize false positives, sequence reduction stably increases all the metrics.

	\begin{table}[t]
		\centering
		\scalebox{0.9}{
			\begin{tabular}
				{cc|cc|cc}
				\toprule[1.0pt]
				& Seq.  & \multicolumn{2}{c|}{YouTube} & \multicolumn{2}{c}{DAVIS} \\
				\cmidrule[0.5pt]{3-4} \cmidrule[0.5pt]{5-6}
				Backbone & Reduce & $\mathcal{AP}$ & $\mathcal{AR}@100$ & $\mathcal{J}$ & $\mathcal{F}$ \\
				\midrule[0.5pt]
				\multirow{2}{*}{ResNet-101} & & 19.3 & \textbf{55.1} & 62.4 & 69.5 \\
				& $\checkmark$ & \textbf{43.8} & 53.2 & \textbf{65.0} & \textbf{71.6} \\
				\midrule[0.5pt]
				\multirow{2}{*}{ResNeXt-101} & & 20.7 & \textbf{58.1} & 64.9 & 70.9 \\
				& $\checkmark$ & \textbf{47.6} & 56.0 & \textbf{67.0} & \textbf{73.8} \\
				\bottomrule[1.0pt]
			\end{tabular}
		}
		\vspace{0.05in}
		\caption{Sequence reduction analysis on YouTube-VIS and DAVIS-UVOS validation set. } 
		\label{tab: abl_nms}
		\vspace{-0.1in}
	\end{table}

	\begin{table}[t]
		\centering
		\scalebox{0.9}{
			\begin{tabular}
				{c|ccc}
				\toprule[1.0pt]
				CA. Reduce & ResNet-50 & ResNet-101 & ResNeXt-101 \\
				\midrule[0.5pt]
				 & 40.4 & 43.8 & 47.6 \\
				$\checkmark$ & \bf{41.5} & \bf{45.1} & \bf{48.3} \\
				\bottomrule[1.0pt]
			\end{tabular}
		}
		\vspace{0.05in}
		\caption{Ablations of Category-Aware Reduction (CA. Reduce) on different backbones. We reports $\mathcal{AP}$ for YouTube-VIS. } 
		\label{tab: abl_ca_reduce}
		\vspace{-0.2in}
	\end{table}

	\vspace{-0.15in}
	\paragraph{Category-Aware Reduction}
	In the evaluation of category-aware metrics (\eg, $\mathcal{AP}$), new redundancy appears after the category assignment. Sequences assigned to the same category conflict with each other in the final evaluation.
	These redundant sequences can be filtered by applying the same sequence reduction techniques (Sec.~\ref{Sec:SPR}) for each category.
	Ablation results on Tab.~\ref{tab: abl_ca_reduce} demonstrate that such a category-aware reduction post-processing stably improves the accuracy on different backbones.

	\begin{figure}[t]
		\begin{center}
			\includegraphics[width=1.0\linewidth]{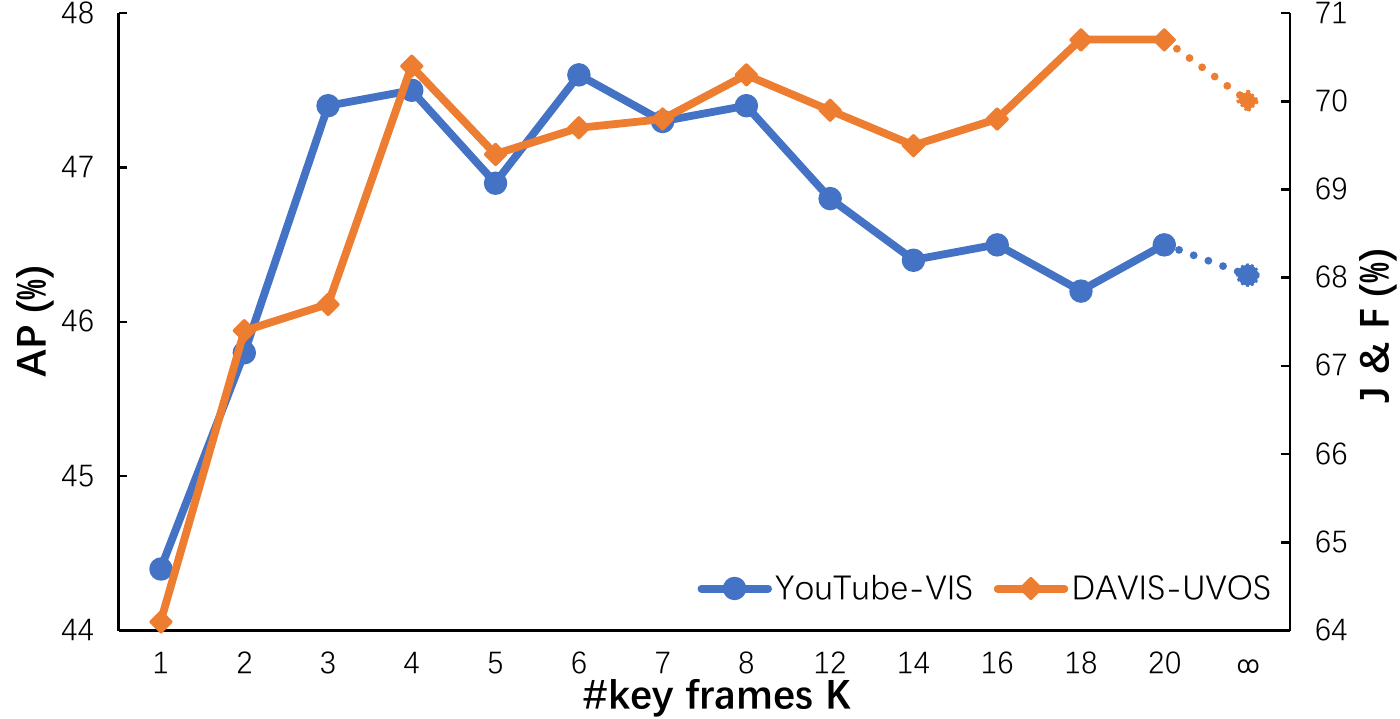}
		\end{center}
		\vspace{-0.15in}
		\caption{Key frame analysis on YouTube-VIS and DAVIS-UVOS validation set. `\#key frames' denotes the number of selected key frames in a video, where $K=\infty$ means all frames are key frames. }
		\label{fig:abl_keynum}
		\vspace{-0.15in}
	\end{figure}
	
	\vspace{-0.2in}
	\paragraph{Key Frames} \label{sec:abl_keyframe}
	In our inference paradigm, the number of key frames (hyper-parameter $K$) plays an important role in controlling the trade-off between accuracy and efficiency. 
	As shown in Fig.~\ref{fig:abl_keynum}, our model performs poorly when $K=1$ since one sequence proposal is sensitive to segmentation quality at the key frame (see also Fig.~\ref{fig:key}) and may miss some instances.
	When sampling more key frames, accuracy increases dramatically as sufficient sampling improves robustness and recall for the same instance. 
	With more key frames sampled, accuracy fluctuates, probably because the quality of instance sequences varies among different key frames and the estimated sequence score (Eq.~(\ref{eq:seq_score})) in NMS cannot effectively reflect the priority of the sequences.
	
	Note that the accuracy in YouTube-VIS gradually drops as $K \ge 8$, since its evaluation metric (Sec. \ref{Sec:dataset}) is sensitive to false positives from residual redundant sequences after NMS.
	We set $K$ to $6$ and $4$ for YouTube-VIS and DAVIS-UVOS by default.
	
	\section{Conclusion}
	In this paper, we have proposed a new paradigm to tackle the task of video instance segmentation, which requires no tracking/matching part to avoid error accumulation and ensures high recall.
	Following the paradigm, we design our framework named Seq Mask R-CNN, which incorporates a newly-designed propagation head on Mask R-CNN.
	The extensive experiments verify the effectiveness of our framework.
	Besides, this work provides a new perspective on the VIS task, which motivates future work on extending image-level methods to video domain.
	
	\clearpage
	{\small
		\bibliographystyle{ieee_fullname}
		\bibliography{egbib}
	}
	
	\newpage
	
	\section*{A. Overview}
	We provide additional details in this supplementary file. Sec.~\textcolor{red}{B} describes the details of the sequence proposals reduction. In Sec.~\textcolor{red}{C}, we describe more details regarding the Seq-Prop head. In Sec.~\textcolor{red}{D}, we clarify more details of the implementation. More discussions are presented in Sec.~\textcolor{red}{E}. More visual results are shown in Sec.~\textcolor{red}{F}.
	
    \section*{B. Sequence Proposals Reduction} \label{sec:seq_reduce}
	With the defined input sequence set $S$, sequence score (Eq.~(1)) and sequences IoU (Eq.~(2)) described in our main paper (Sec.~3.2), we apply the traditional NMS algorithm on the sequence set to reduce the redundant sequences. The algorithm for sequence proposal reduction is illustrated in Alg.~\ref{alg:iou}. The IoU threshold $\theta$ is set to $0.5$ in our experiments.

	\begin{figure*}[h]
		\begin{center}
			\includegraphics[width=1.0\linewidth]{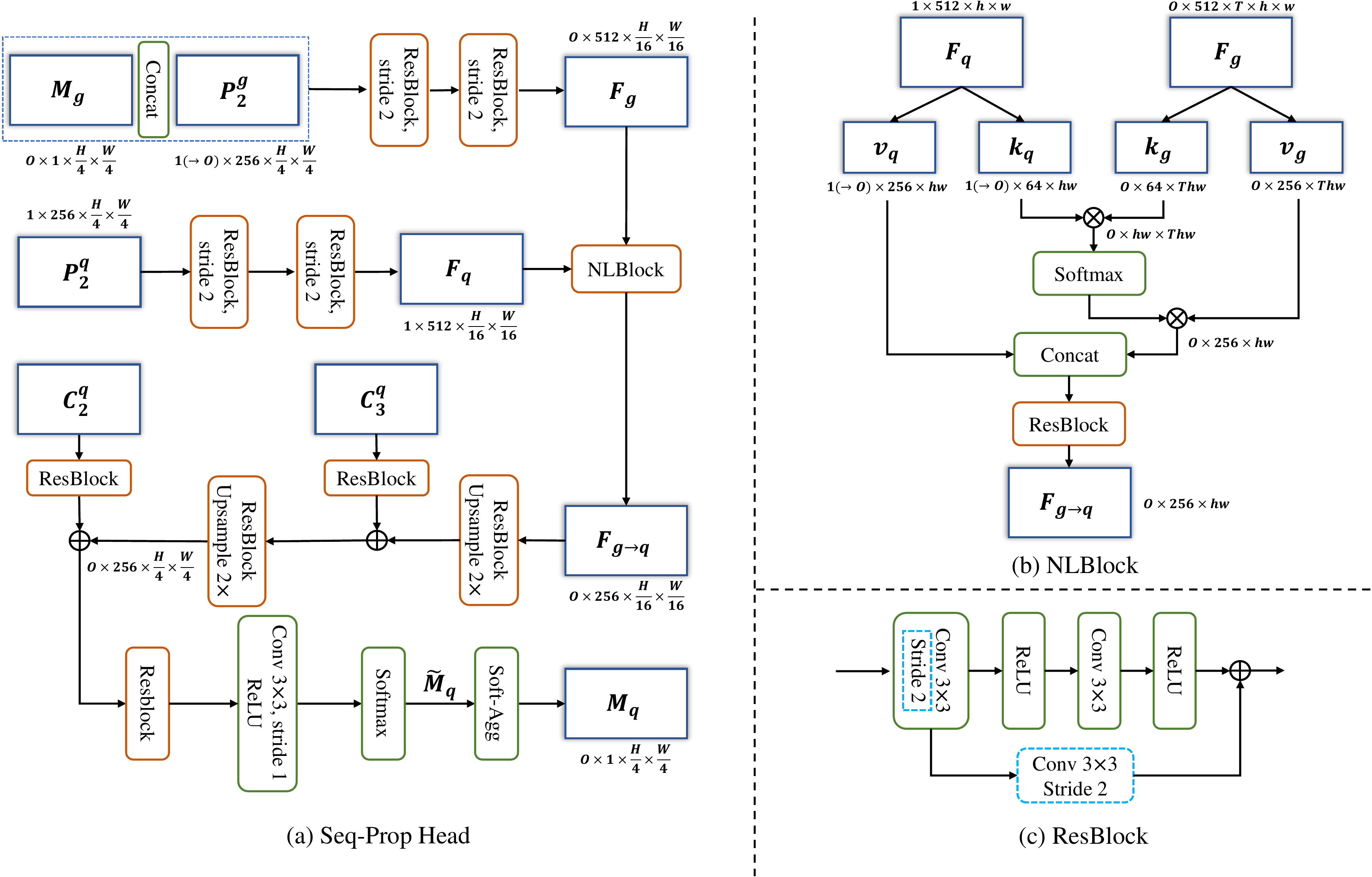}
		\end{center}
		\caption{Architectures of the (a) \textbf{Seq-Prop head}, including the (b) NLBlock ~\cite{wang2018non} and the (c) ResBlock~\cite{he2016deep}. $O$, $T$, $H$ and $W$ indicate instance number, frame number, height and width respectively. `($\to O$)' denotes expanding the tensor along the specific dimension. The `Soft-Agg' operation refers to Eq.~(\ref{eq:softagg}).}
		\label{fig:arc}
	\end{figure*}
	
	\begin{figure*}[t!]
		\begin{center}
			\includegraphics[width=1.0\linewidth]{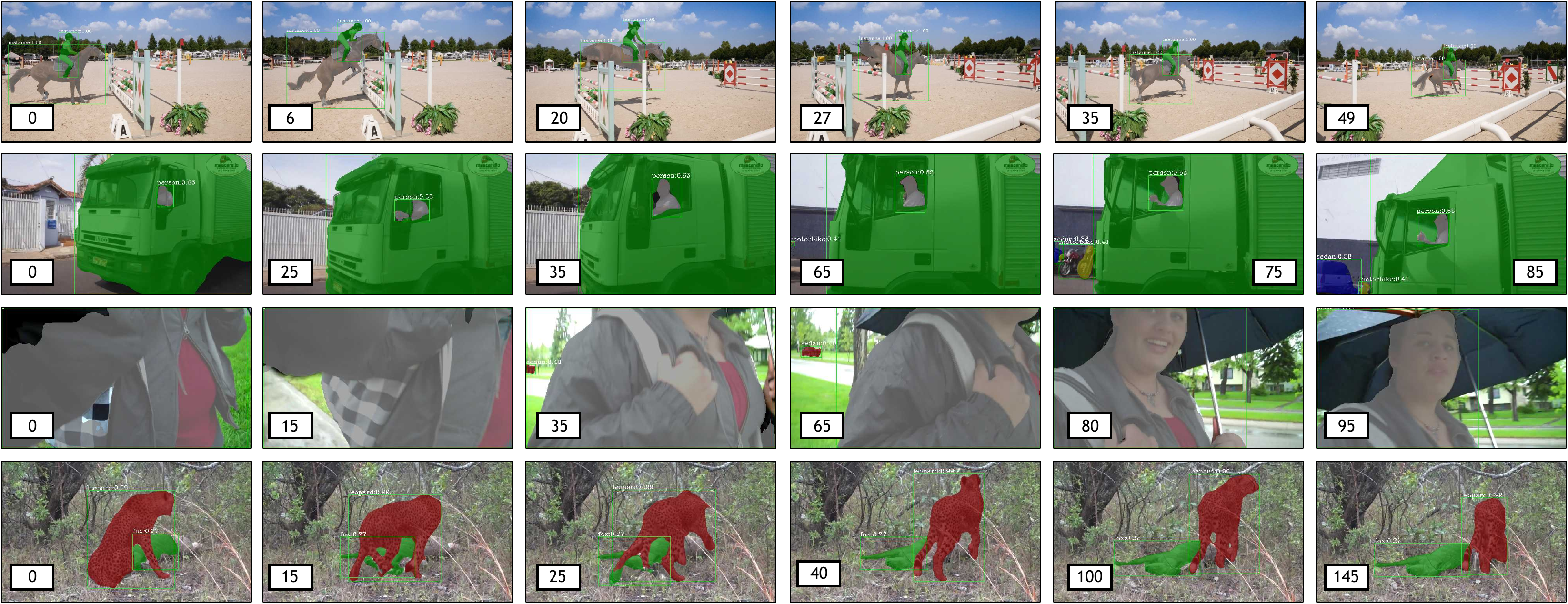}
		\end{center}
		\caption{Visual results in various scenarios on DAVIS-UVOS and YouTube-VIS validation set. Category `Instance' in DAVIS-UVOS denotes the salient generic object. The last row is a failure case.
		Zoom in for details.}
		\label{fig:visu}
		\vspace{0.1in}
	\end{figure*}
	
	\section*{C. Seq-Prop Head} \label{sec:head}
	\vspace{-0.05in}
	\paragraph{Architecture} 
    The detailed architecture of the Seq-Prop head is shown in Fig.~\ref{fig:arc}.
	\vspace{-0.2in}		
	\paragraph{Soft-Agg}
	The soft aggregation \cite{wug2018fast} of estimated mask $\tilde{M}_q(p)$ is defined as 
	\begin{equation}\label{eq:softagg}
	M_q^o(p) = \frac{\tilde{M}_q^o(p)/(1-\tilde{M}_q^o(p))}{\sum_{i=0}^{O}{\tilde{M}_q^i(p)/(1-\tilde{M}_q^i(p))}},
	\end{equation}
	where $\tilde{M}_q^0(p)=\prod_{i=1}^{O}(1-\tilde{M}_q^i(p))$ denotes the background prediction.
    \vspace{-0.05in}
	\paragraph{Training Loss}
	With $\hat{M_q}$ and $M_q$ denoting the ground-truth and predicted masks, the scale-balanced soft IoU loss \cite{lin2019agss} is defined as 
	\begin{equation}
	\begin{footnotesize}
	\begin{aligned}
	& \mathcal{L}(\hat{M_q}, M_q) = 1 - \frac{1}{O}\sum_{o=1}^{O}{\frac{\sum_{p}{min(\hat{M}_q^o(p), M_q^o(p))}}
		{\sum_{p}{max(\hat{M}_q^o(p),M_q^o(p)))}}} ,
	\end{aligned}
	\end{footnotesize}
	\label{equ:iouloss}
	\end{equation}
	
	where $M_q^o(p)$ denotes the value of the $o^{th}$ instance in query mask $M_q$ at pixel $p$ and so as $\hat{M}_q^o(p)$.

	\begin{algorithm}[t]\label{alg:iou} 
		\caption{Sequence Proposals Reduction} 
		\KwIn{Input sequence set $S=\{S_k^o\}$; \\
			~~~~~~~~~~~~Its classification score $C(S)=\{C(S_k^o)\}$; \\
			~~~~~~~~~~~~Its mask sequence set $M(S)=\{M(S_k^o)\}$; \\
			~~~~~~~~~~~~where $k = 0,1,...,K-1$ and \\ 
			~~~~~~~~~~~~~~~~~~~~~~~$o = 0,1,...,O-1$. \\
			~~~~~~~~~~~~IoU threshold $\theta$. }
		\KwOut{Final sequence set $\widehat{S}$}
		
		$\widehat{S} \leftarrow \{\}$\;
		\While{$S \neq \varnothing$}
		{
			$(k', o') \leftarrow$ argmax $C(S)$\;
			$V \leftarrow S_{k'}^{o'}$ \;
			$\widehat{S} \leftarrow \widehat{S} \cup V; S \leftarrow S - V$\;
			\For{$S_k^o$ in $S$}
			{
				\If{IoU($M(V)$, $M(S_k^o)$) $\ge \theta$}
				{
					$S \leftarrow S - S_k^o$\;
					$C(S) \leftarrow C(S) - C(S_k^o)$\;
					$M(S) \leftarrow M(S) - M(S_k^o)$\;
				}
			}
		}
		return $\widehat{S}$\;	
	\end{algorithm}
	
	\vspace{-0.05in}
	\section*{D. Implementation Details} \label{sec:implement}
	\paragraph{Training} We follow the training setup as in \cite{he2017mask}. We train our model for $4$ epochs in the main-training stage and $5$ epochs for the finetuning stage. In the main-training stage, we adopt the SGD optimizer with an initial learning rate of $5e$-$3$. The learning rate decays by a factor of $10$ at the $3$ and $4$ epochs. In the finetuning stage, the learning rate is fixed at $5e$-$5$. The batch size is set to the maximum possible magnitude for different backbones. Our model is initialized with the pre-trained weight of Mask R-CNN \cite{he2017mask} on COCO, while the additional propagation head is initialized randomly.
	
	\vspace{-0.2in}
	\paragraph{Inference} During the testing stage, RPN generates $200$ proposals for each key frame. For a key frame, the detected instances are sorted by score and the top $10$ (\ie, $O$) ones with scores higher than $0.2$ are used for generating sequence proposals. 
	The memory pool is updated every 5 frames in the Seq-Prop head.
	
	\section*{E. Discussion} \label{sec:discuss}
	\paragraph{Comparison with MaskProp} 
	Since MaskProp \cite{bertasius2020maskprop} does not release codes or pre-computed results, a qualitative comparison is infeasible.
	Nevertheless, Tab.~\ref{tab:exp_ytv} can give some hints about the difference between MaskProp and our method. In Tab.~1, when the $\mathcal{AP}$ is close ($47.6$ $\vs$ $46.6$), our method has a better $\mathcal{AR}@10$ than MaskProp ($56.0$ $\vs$ $52.6$). It indicates our method has more true positives in the top-10 scoring instances. However, with higher recall, they have similar $\mathcal{AP}$. This suggests that MaskProp has the better scoring (according to the rules of mAP), which may be because of the stronger backbone employed (i.e., STSN-ResNeXt-101). When using the same backbone (ResNeXt-101), our method is better than MaskProp by $3.3\%$ in terms of $\mathcal{AP}$.

	\begin{figure}[h]
		\begin{center}
			\includegraphics[width=1.0\linewidth]{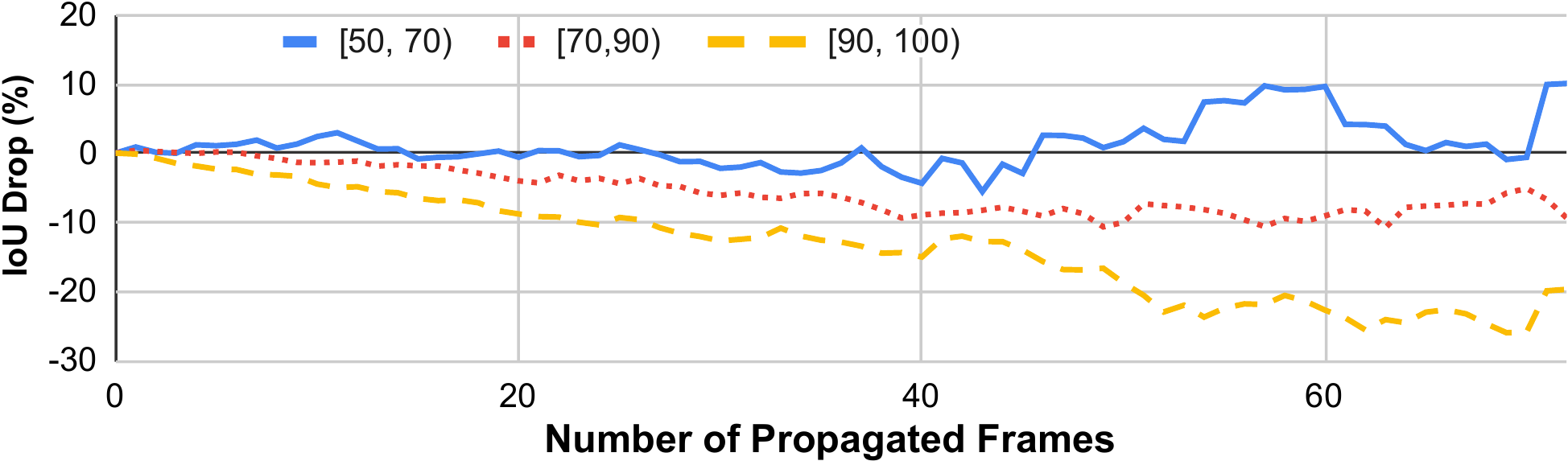}
		\end{center}
		\vspace{-0.15in}
		\caption{IoU (\ie, $\mathcal{J}$-Mean) drops regarding propagated distance on DAVIS with ResNeXt-101. 
			`$[a,b)$' indicates the group of sequence proposals where the IoU between the initial mask and corresponding ground-truth is in the range of $[a\%, b\%)$.
		}
		\label{fig:abl_proplen}
		\vspace{-0.2in}
	\end{figure}
	
	\paragraph{Distant Frame Pairs}
	During the inference stage, the Seq-Prop head propagates segmented masks (in key frames) to other frames. It is worth investigating the effectiveness of propagating to distant frames.
	To this end, we group sequence proposals into three types with different initial quality (\ie, IoU) at the starting key frame (Fig.~\ref{fig:abl_proplen}).
	The IoU of propagated masks drops by around $20\%$ with high-quality initial masks. 
	As the initial mask quality lowers, the IoU at distant frames drops less and even rises. This may be due to the reason that the propagation head learned shape information for objects.

	\vspace{-0.07in}
	\section*{F. Visual Results} \label{sec:visu}
	\vspace{-0.05in}
	We provide more visual results in Fig.~\ref{fig:visu}. The last row is a failure case, where the deer is misclassified as a `fox'.
	
\end{document}